\documentclass{article}

\usepackage{arxiv}

\usepackage[utf8]{inputenc} 
\usepackage[T1]{fontenc}    
\usepackage{hyperref}       
\usepackage{url}            
\usepackage{booktabs}       
\usepackage{amsfonts}       
\usepackage{nicefrac}       
\usepackage{microtype}      
\usepackage{lipsum}
\usepackage{graphicx}
\usepackage{multirow, array}
\usepackage{amsmath}
\usepackage{amssymb}
\graphicspath{ {./images/} }

\title{A Two-Stage Data Association Approach for 3D Multi-Object Tracking}

\author{
 Minh-Quan Dao, Vincent Frémont \\
  Laboratoire des Sciences du Numérique de Nantes (LS2N)\\
  École Centrale de Nantes\\
  Nantes, 44300 \\
  \texttt{minh-quan.dao@ls2n.fr, vincent.fremont@ls2n.com} \\
  
}

\begin{document}
\maketitle
\begin{abstract}
Multi-object tracking (MOT) is an integral part of any autonomous driving pipelines because it produces trajectories which has been taken by other moving objects in the scene and helps predict their future motion. Thanks to the recent advances in 3D object detection enabled by deep learning, track-by-detection has become the dominant paradigm in 3D MOT. In this paradigm, a MOT system is essentially made of an object detector and a data association algorithm which establishes track-to-detection correspondence. While 3D object detection has been actively researched, association algorithms for 3D MOT seem to settle at a bipartie matching formulated as a linear assignment problem (LAP) and solved by the Hungarian algorithm. In this paper, we adapt a two-stage data association method which was successful in image-based tracking to the 3D setting, thus providing an alternative for data association for 3D MOT. Our method outperforms the baseline using one-stage bipartie matching for data association by achieving 0.587 AMOTA in NuScenes validation set.
\end{abstract}


\section{Introduction}
Multi-object tracking have been a long standing problem in computer vision and robotics community since it is a crucial part of autonomous systems. From the early work of tracking with hand-craft features, the revolution of deep learning which results in highly accurate object detection models \cite{ren2016faster,liu2016ssd,redmon2016you} has shifted the focus of the field to the track-by-detection paradigm \cite{bewley2016simple, scheidegger2018mono}. In the framework of this paradigm, tracking algorithms receive a set of object detection, usually in the form of bounding boxes, at each time step and they aim to link detection of the same object across time to form trajectories. 

While image-based methods of this paradigm have reached a certain maturity, 3D tracking is still in its early phase where most of the published approaches are originated from successful 2D exemplars. The most popular attempt at 3D tracking with established 2D tracking method is \cite{weng2020ab3dmot} which is an extension of \cite{bewley2016simple} into 3D space. In these works, the tracking algorithm is made of the Hungarian algorithm \cite{kuhn1955hungarian} and Kalman filter. While the former finds track-to-measurement correspondences by solving a linear assignment problem, the later performs prediction and correction of tracks' state. An improvement of \cite{weng2020ab3dmot} is proposed by \cite{chiu2020probabilistic} which replaces the 3D IoU \cite{zhou2019iou} with the Mahalanobis distance as the cost function for the assignment problem. The idea of handling tracking as a matching problem is also used in the context of end-to-end learning \cite{liang2020pnpnet,yin2020center,luo2018fast}. \cite{liang2020pnpnet} solves the tracking task in same fashion as \cite{weng2020ab3dmot}; however, this work trains a sub network for calculating the cost function of the assignment problem and the correction step is carried out by another sub network instead of the Kalman filter.\cite{yin2020center,luo2018fast}
train deep models to predict tracks position in the following frames along with generating detection and the track-to-detection correspondences are found by greedy matching.

Even though 3D tracking has been progressed rapidly thanks to the availability of standardized large scale benchmarks such as KITTI \cite{geiger2013vision}, NuScenes \cite{caesar2020nuscenes}, Waymo Open Dataset \cite{sun2020scalability}, the focus of the field is placed on developing better object detection models rather than developing better tracking algorithm as evidenced in the Table.\ref{tab:agg-tracking-methods}. There are two trends can be observed in this table. First, tracking performance experiences significant boost when a better object detection model is introduced. Second, the method of AB3DMOT \cite{weng2020ab3dmot} is favored by most recent 3D tracking systems.

\begin{table}[h]
    \caption{Summary of tracking methods which details are published in the leader board of NuScenes and Waymo Open Dataset}
    \label{tab:agg-tracking-methods}

\begin{tabular}{ |m{1.5cm}|m{2.5cm}|m{5cm}|m{1.25cm}|m{2.25cm}|m{1cm}| } 
\hline
Dataset & Method Name & Tracking Method & AMOTA & Object Detector & mAP \\
\hline
\multirow{5}{1.5cm}{NuScenes} & CenterPoint \cite{yin2020center} & Greedy closest-point matching & 0.650 & CenterPoint & 0.603 \\
\cline{2-6}
& PMBM* & Poisson Multi-Bernoulli Mixture filter \cite{pmbm} & 0.626 & CenterPoint & 0.603 \\
\cline{2-6}
& StanfordIPRL-TRI \cite{chiu2020probabilistic} & Hungarian algorithm with Mahalanobis distance as cost function and Kalman Filter & 0.550 & MEGVII \cite{zhu2019megvii} & 0.519 \\ 
\cline{2-6}
& AB3DMOT \cite{weng2020ab3dmot} & Hungarian algorithm with 3D IoU as cost function and Kalman Filter & 0.151 & MEGVII & 0.519 \\
\cline{2-6}
& CenterTrack & Greedy closest-point mathcing & 0.108 & CenterNet \cite{zhou2019objects} & 0.388 \\
\hline
\multirow{4}{1.5cm}{Waymo} & HorizonMOT \cite{ding20201st} & 3-stage data associate, each stage is an assignment problem solved by Hungarian algorithm & 0.6345 & AFDet \cite{ge2020afdet} & 0.7711 \\
\cline{2-6}
& CenterPoint & Greedy closest-point matching & 0.5867 & CenterPoint & 0.7193 \\
\cline{2-6}
& PV-RCNN-KF & Hungarian algorithm and Kalman Filter & 0.5553 & PV-RCNN \cite{shi2020pv} & 0.7152 \\ 
\cline{2-6}
& PPBA AB3DMOT & Hungarian algorithm with 3D IoU as cost function and Kalman Filter & 0.2914 & PointPillars and PPBA\cite{cheng2020improving} & 0.3530 \\
\hline
\end{tabular}
\end{table}

The reason of AB3DMOT's popularity is that despite its simplicity, it achieves competitive result in challenging datasets at significantly high frame rate (more than 200 FPS). However, such simplicity comes at the cost of the MOT system being vulnerable to false associations due to occlusion or imperfect detections which is case for objects in a clutter or far away from the ego vehicle. 

Aware of the shortage of a generic 3D tracking algorithm which can better handle occlusion and imperfect detections so that to limit the false track-to-detection correspondence, yet remains relatively simple, we adapt the image-based tracking method proposed by \cite{bae2014robust} to the 3D setting. Specifically, this method is a two-stage data association scheme. In this scheme, each tracked trajectory is called a tracklet and is assigned a confidence score computed based on how well associated detection matches with tracklet. The first association stage aims to establish the correspondence between high confident tracklets and detection. The second stage matches the left over detection with the low confident tracklets as well as link low confident tracklets to high confident ones if they meet a certain criterion. 

In this paper, we make two contributions
\begin{itemize}
    \item Our main contribution is the adaptation of an image-based tracking method to the 3D setting. In details, we exploit a kinematically feasible motion model, which is unavailable in 2D, to facilitate objects pose prediction. This model in turn defines the minimal state vector needed to be tracked.
    \item Extensive experiment carried out in various datasets proves the effectiveness of our approach. In fact, our better performance, compared to AB3DMOT-style models, show that adding a certain degree of re-identification can improve the tracking performance while keeping the added complexity to the minimum.
\end{itemize}

\section{Related work}
A multi-object tracking system in the track-by-detection paradigm consists of an object detection model, a data association algorithm and a filtering method. While the last two components are domain agnostic, object detection models, especially learning-based methods, are tailored to their operation domain (e.g images or point clouds). This paper targets autonomous driving where objects pose are required thus interest in 3D object detection models. However, developing such a model is not in the scope of this paper, instead we use the detection result provided by baseline models of benchmarks (e.g. PointPillars of NuScenes) to focus on the data association algorithm and to have a fair comparison. Interested readers are referred to \cite{arnold2019survey} for a review of 3D object detection.     

Data association via the Hungarian algorithm was early explored in \cite{geiger2013traffic} where a 2-stage tracking scheme was proposed for offline 2D tracking. Firstly, detections are linked frame-by-frame to form tracklets. The affinity matrix of the Hungarian algorithm is established by geometric and appearance cue. While the geometric cue is the 2D Intersection over Union (IoU), the appearance cue is the correlation between two bounding boxes. Secondly, tracklets are associated to each other to compensate trajectory fragments and ID switch due to occlusion. This association is also carried out the Hungarian algorithm.

Due to its batch-processing nature \cite{geiger2013traffic} cannot be applied to online tracking, \cite{bewley2016simple} overcomes this by eliminating the second stage and let objects which temporarily left the sensor's field of view reenter with new IDs. Despite its simplicity, SORT - the method proposed by \cite{bewley2016simple} achieve competitive result in MOT15\cite{leal2015motchallenge} with lightning-fast inference speed (260 Hz). The success of SORT inspired \cite{weng2020ab3dmot} to adapt it to 3D setting by using 3D IoU as the affinity function. The performance of SORT in 3D setting is later improved in \cite{chiu2020probabilistic} which shows the use of Mahalanobis distance is superior to 3D IoU. \cite{mauri2020deep} integrates the 3D version of SORT into a complete perception pipeline for autonomous vehicles.

The two-stage association scheme is adapted to online tracking in \cite{bae2014robust} which proposes a confidence score to quantify tracklets quality. Based on this score, tracklets are associated with detections or another tracklets, or terminated. The appearance model learned by ILDA in \cite{bae2014robust} is improved by deep learning in the follow-up work \cite{bae2017confidence}. Recently, this association scheme is revisited in the context of image-based pedestrian tracking by \cite{yang2019efficient} which proposed to use the rank of the Hankel matrix as tracklets motion affinity. 

Differ from \cite{bae2014robust} and its related works, this paper applies the two-stage association scheme to online 3D tracking. In addition, we can provide competitive result despite relying solely on geometric cue to compute tracklet affinity thanks to the Constant Turning Rate and Velocity (CTRV) motion model which can accurately predict objects position in 3D space by exploiting their kinematic.

\section{Method} \label{sec:method}
\subsection{Problem Formulation}
Online multi-object tracking (MOT) in the sense of track-by-detection aims to gradually grow the set of tracklets $\mathbb{T}_{0:t} = \{\mathcal{T}^i\}_{i = 1}^{|\mathbb{T}_{0:t}|}$ by establishing correspondences with the set of detections received at every time step $\mathbb{D}_t = \{d_{t}^j\}_{j = 1}^{|\mathbb{D}|}$ and updating tracklets state accordingly. A tracklet is a collection of state vectors corresponding to the same object $\mathcal{T}^i = \{\mathrm{x}_k^i | 0 \leq t_s^i \leq k \leq t_e^i \leq t \}$, here $t_s^i, t_e^i$ are respectively the starting- and ending-time of the tracklet. A detection $d_{t}^j$ at time step $t$ encapsulates information of a 3D bounding box including the position of its center in a common reference frame $[x, y, z]$, heading $\theta$, and size $[w, l, h]$. It is worth to notice that in the context of autonomous driving, objects are assumed to remain in contact with the ground; therefore, their detections are up-right bounding boxes which orientation is described by a single number - the heading angle.

The correspondence between $\mathbb{T}_{0:t}$ and $\mathbb{D}_t$ can be formally defined as finding the set $\mathbb{T}_{0:t}^*$ that maximizes its likelihood given $\mathbb{D}_t$.

\begin{equation}
    \mathbb{T}_{0:t}^* = \underset{\mathbb{T}_{0:t}}{\mathsf{argmax}} \text{  } \mathtt{p}\left(\mathbb{T}_{0:t} | \mathbb{D}_t \right)
    \label{eq:correspondece_as_MAP}
\end{equation}

Due the exponential growth of possible associations between $\mathbb{T}_{0:t}$ and $\mathbb{D}_{t}$, Equation.\eqref{eq:correspondece_as_MAP} is computationally intractable after a few time steps. In this paper, such a correspondence is approximated by the two-stage data association proposed by \cite{bae2014robust} as shown in the following.

\subsection{Two-stage Data Association}
\subsubsection{Tracklet Confidence Score}
The reliability of a tracklet is quantified by a confidence score which is calculated based on how well associated detections match with its states across its life span and how long its corresponding object was undetected.

\begin{equation}
    \mathsf{conf}\left(\mathcal{T}^i\right) = \left(\frac{1}{L^i}\sum_{k \in [t_s^i, t_e^i]| v^i(k) = 1} \Lambda\left(\mathcal{T}^i, d_k^j\right) \right) \times \exp \left(-\beta \frac{W}{L_i}\right)
    \label{eq:tracklet_conf}
\end{equation}
where $v^i(k)$ is a binary indicator which takes 1 if the tracklet has a detection $d_k^j$ associated with it at time step $k$, and 0 otherwise. $L^i$ is the number of time step that the traklet gets associated with a detection. $\Lambda (\cdot)$ is the affinity function which detail will be presented later. $\beta$ is a tuning parameter which takes high value if the object detection model is accurate. $W = t - t_s^i - L^i + 1$ is the number of time step that tracklet was undetected (i.e. did not have associated detection) calculated from its birth to the current time step $t$.

\begin{figure}[htb]
\includegraphics[width=\linewidth]{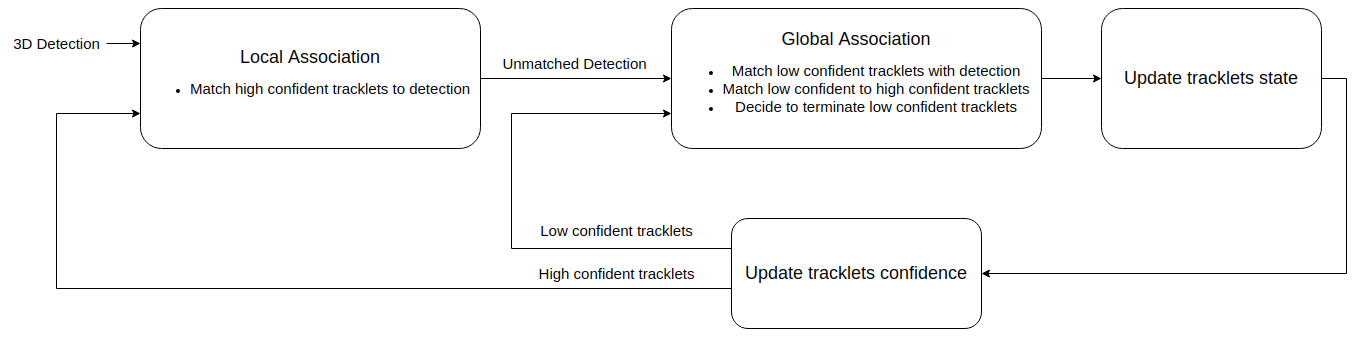}
\caption{The pipeline of two-stage data association. The first stage - local association establish the correspondences between detections at this time step $\mathbb{D}_{t}$ and high confidence tracklets $\mathbb{T}_{0:t}^{h}$. Then, global association stage matches each low confidence tracklets $\mathcal{T}^{i, l}$ with either a high confidence tracklet or a left-over detection, or terminates it.}
\label{fig:two-stage-pipeline}
\end{figure}

Applying a threshold $\tau^c$ this confidence score divides the set $\mathbb{T}_{0:t}$ into a subset of high confidence tracklets $\mathbb{T}_{0:t}^{h} = \{\mathcal{T}^{i, h} | \mathsf{conf}\left(\mathcal{T}^i\right) > \tau^c\}$ and a subset of low confidence tracklets $\mathbb{T}_{0:t}^{l} = \{\mathcal{T}^{i, l} | \mathsf{conf}\left(\mathcal{T}^i\right) \leq \tau^c\}$. These two subsets are the fundamental elements of the two-stage association pipeline showed in Figure.\ref{fig:two-stage-pipeline}

\subsubsection{Local Association}
In this association stage, high confident tracklets are extended by their correspondence in the set of detections $\mathbb{D}_t$. This tracklet-to-detection is found by solving the a linear assignment problem (LAP) characterized by the cost matrix $\mathbf{L}$ as following

\begin{equation}
    \mathbf{L} = \begin{bmatrix}
        l_{i, j}
    \end{bmatrix} \in \mathbb{R}^{h \times d}, \text{  with } l_{i, j} = -\Lambda \left(\mathcal{T}^i, d_t^j \right)
\end{equation}
where $h, d$ are respectively the number of high confidence tracklets and the number of detections. The intuition of this association stage is that because tracklets with high confidence have been tracked accurately for a number of time steps, the affinity function can identify if a detection is belong to the same object as the tracklet with high accuracy, thus limiting the possibility of false correspondences. In addition, low confidence tracklets are usually resulted from fragment trajectories or noisy detections, excluding them from this association stage help reduces the ambiguity.

\subsubsection{Global Association}
As shown in Figure.\ref{fig:two-stage-pipeline}, the global assocaition stage carries out the following tasks
\begin{itemize}
    \item Matching low confidence tracklets with high confidence ones
    \item Matching low confidence tracklets with detections left over by the local association stage
    \item Deciding to terminate low confidence tracklets
\end{itemize}
These tasks are simultaneously solved as a LAP formulated by the following cost matrix

\begin{equation}
    \mathbf{G}_{(l + d') \times (h + l)} = \begin{bmatrix}
        \mathbf{A}_{l \times h} & B_{l \times l} \\
        \infty_{d' \times h} & C_{d' \times l}
    \end{bmatrix}
    \label{eq:global_cost}
\end{equation}
here, $l, d$ are respectively the number low confidence tracklets and detections left over by the local association stage. Recall $h$ is the number of high confidence tracklets. Submatrix $\mathbf{A}$ is the cost matrix of the event where low confidence tracklets are matched with high confidence ones 

\begin{equation}
    \mathbf{A} = [a_{i, j}] \in \mathbb{R}^{l \times h}, \text{ with } a_{i, j} = -\Lambda\left(T^{i, l}, T^{j, h}\right)
\end{equation}
Submatrix $\mathbf{B}$ represents the event where low confidence tracklets are terminated.

\begin{equation}
    \mathbf{B} = [b_{i, j}] \in \mathbb{R}^{l \times l}, \text{ with }  b_{i, j} = \begin{cases}
    -\log \left(1 -  \mathsf{conf}\left(\mathcal{T}^i\right) \right), \text{  if } i = j \\
    \infty, \text{  otherwise}
    \end{cases}
\end{equation}
Finally, submatrix $\mathbf{C}$ is the cost of the associating low confidence tracklets with detections left over by local association stage.

\begin{equation}
    \mathbf{C} = [c_{i, j}] \in \mathbb{R}^{d' \times l}, \text{ with } c_{i, j} = -\Lambda\left(\mathcal{T}^j, d^i_t\right)
\end{equation}

The solution to the LAP in this stage and in the Local Association stage is the association that \textit{minimize} the cost and can be either found by the Hungarian algorithm for the optimal solution or by a greedy algorithm which interatively pick and remove correspondence pair with the smallest cost until there is no pair has cost less than a threshold (the detail of this greedy algorithm can be found in \cite{chiu2020probabilistic}). 

Once a detection is assocatied with a tracklet, its position and heading is used to update the tracklet's state according to the equation of the Kalman Filter, while its sizes is averaged with tracklet's sizes in past few time steps to result in updated sizes. Detections do not get associated in the global association stage are used to initialize new tracklets.

\subsubsection{Affinity Function}
Affinity function $\Lambda (\cdot)$ is to compute how similar a detection to a tracklet or a tracklet to another. As mentioned earlier, due to the lack of colorful texture in point cloud, the affinity function used in this work is just comprised of geometric cue. Specifically, it is the sum of position affinity and size affinity.

\begin{equation}
    \Lambda (\mathcal{T}^i, d_t^j) = \Lambda (\mathcal{T}^i, d_t^j)^p + \Lambda (\mathcal{T}^i, d_t^j)^s
    \label{eq:affinity_general}
\end{equation}

\begin{figure}[htb]
\includegraphics[width=\linewidth]{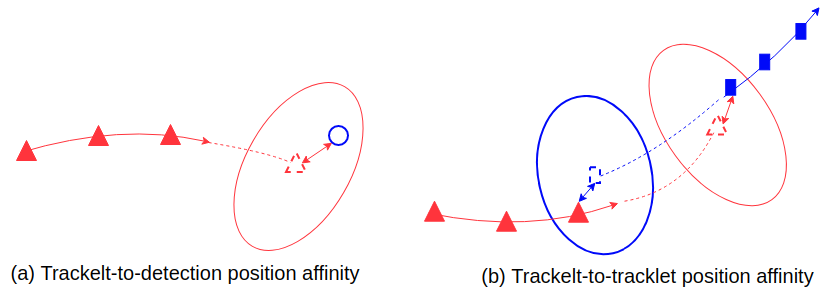}
\caption{The computational scheme of position affinity. The filled triangles (or rectangles) are subsequent states of a tracklet. The colored arrow represents the time order: the closer to the tip, the more recent the state. The triangle (or rectangle) in dash line is the state propagated forward (or backward) in time. The covariance of these propagated states are denoted by ellipses with the same color. The two-headed arrows indicate the Mahalanobis distance. In the subfigure (a), the blue circle denotes a detection.}
\label{fig:position_aff_comp}
\end{figure}

The scheme for computing position affinity between a tracklet and a detection or between two tracklets are shown in Figure.\ref{fig:position_aff_comp}.

As shown in Figure.\ref{fig:position_aff_comp}.a, the position affinity $\Lambda (\mathcal{T}^i, d_t^j)^p$ between a tracklet $\mathcal{T}^i$ and a detection $d_t^j$ is defined as the Mahalanobis distance between tracklet's last state propagated to the current time step $t$ and the measurement vector $\mathrm{z}_t^j$ extracted from $d_t^j$

\begin{equation}
    \Lambda (\mathcal{T}^i, d_t^j)^p = \left( \mathsf{h} \left( \Bar{\mathrm{x}}^i_e \right) - \mathrm{z}_t^j\right)^T \cdot \mathrm{S}^{-1} \cdot \left( \mathsf{h} \left( \Bar{\mathrm{x}}^i_e \right) - \mathrm{z}_t^j\right)
    \label{eq:tracklet-position-aff-t2d}
\end{equation}
where $\Bar{\mathrm{x}}^i_e$ is last state of tracklet $\mathcal{T}^i$ propagated to the current time step using the motion model which will be presented below. $\mathsf{h}(\cdot)$ is the measurement model computing the expected measurement using the inputted state and the measurement vector $\mathrm{z}_t^j = [x, y, z, \theta]^T$. The matrix $\mathcal{S}$ is the covariance matrix of the innovation (i.e. the difference between expected measurement $\mathsf{h} \left( \Bar{\mathrm{x}}^i_e \right)$ and its true value $\mathrm{z}_t^j$)

\begin{equation}
    \mathbf{S} = \mathbf{H} \cdot \mathbf{P} \cdot \mathbf{H}^T + \mathbf{R}
\end{equation}
here, $\mathbf{H} = \delta \mathsf{h} / \delta \mathrm{x}$ is the Jacobian of the measurement model, $\mathbf{P}, \mathbf{R}$ are covariance matrix of $\Bar{\mathrm{x}}^i_e$ and $\mathrm{z}_t^j$, respectively.

In the case of two tracklets $\mathcal{T}^i$ and $\mathcal{T}^j$, assuming $\mathcal{T}^j$ starts after $\mathcal{T}^i$ ended, their motion affinity is, according to Figure.\ref{fig:position_aff_comp}.b, is the sum of

\begin{itemize}
    \item Mahalanobis distance between the last state of $\mathcal{T}^i$ propagated forward in time and the first state of $\mathcal{T}^j$
    \item Mahalanobis distance between the first state of $\mathcal{T}^j$ propagated backward in time and the last state of $\mathcal{T}^i$
\end{itemize}

\begin{equation}
    \Lambda (\mathcal{T}^i, \mathcal{T}^j)^p = \Lambda (\mathcal{T}^j, \Bar{\mathrm{x}}^i_e)^p + \Lambda (\mathcal{T}^i, \Bar{\mathrm{x}}^j_s)^p
    \label{eq:tracklet-position-aff-t2t}
\end{equation}
here, $\Bar{\mathrm{x}}^i_e$ is the last state of tracklet $\mathcal{T}^i$ propagated forward in time to the first time step of tracklet $\mathcal{T}^j$, while $\Bar{\mathrm{x}}^j_s$ is the first state of tracklet $\mathcal{T}^j$ propagated backward in time to the last time step of tracklet $\mathcal{T}^i$. 

The size affinity $\Lambda (\mathcal{T}^i, d_t^j)^s$ is computed as following

\begin{equation}
    \Lambda \left(\mathcal{T}^i, d_t^j\right)^s = -\frac{|w_e^i - w_t^j|}{w_e^i + w_t^j} \cdot \frac{|l_e^i - l_t^j|}{l_e^i + l_t^j} \cdot \frac{|h_e^i - h_t^j|}{h_e^i + h_t^j}
\end{equation}
here, $[w_e^i, l_e^i, h_e^i]$ are the size of the last state of tracklet $\mathcal{T}^i$, while $[w_t^j, l_t^j, h_t^j]$ are the size of the detection $d_t^j$. In the case of two tracklets $\mathcal{T}^i$ and $\mathcal{T}^j$, assuming $\mathcal{T}^j$ starts after $\mathcal{T}^i$ ended, there size affinity is

\begin{equation}
    \Lambda \left(\mathcal{T}^i, \mathcal{T}^j\right)^s = -\frac{|w_e^i - w_s^j|}{w_e^i + w_s^j} \cdot \frac{|l_e^i - l_s^j|}{l_e^i + l_s^j} \cdot \frac{|h_e^i - h_s^j|}{h_e^i + h_s^j}
    \label{eq:tracklet-size-aff}
\end{equation}
The subscript $e, s$ in Equation.\eqref{eq:tracklet-size-aff} respectively denote the ending and starting state of a tracklet.

\subsection{Motion Model and State Vector}
Exploiting the fact that objects are tracked in 3D space of a common static reference frame which can be referred to as the world frame, motion of objects can be described by more kinematically accurate models, compared to the commonly used Constant Velocity (CV) model. In this work, we use the Constant Turning Rate and Velocity (CTRV) model to predict motion of car-like vehicles (e.g. cars, buses, trucks), while keep the CV model for pedestrians. 

For a car-like vehicles, its state can be described by 

\begin{equation}
    \mathrm{x} = [x, y, z, \theta, \mathtt{v}, \Dot{\theta}, \Dot{z}]^T
\end{equation}
here, $[x, y, z]$ is the location in the world frame of the center of the bounding box represented by the state vector, $\theta$ is the heading angle, $\mathtt{v}$ is longitudal velocity (i.e. velocity along the heading direction), $\Dot{\theta}, \Dot{z}$ are respectively velocity of $\theta$ and $z$.

The motion on x-y plane of car-like vehicles can be predicted using CTRV as following

\begin{equation}
    \mathrm{x}_{t+1} = \mathrm{x}_t + \begin{bmatrix}
        \frac{\mathtt{v}}{\Dot{\theta}} \left(\sin(\theta + \Dot{\theta}\Delta t) - \sin(\theta) \right) \\
        \frac{\mathtt{v}}{\Dot{\theta}} \left(-\cos(\theta + \Dot{\theta}\Delta t) + \cos(\theta) \right) \\
        \Dot{z} \Delta t \\
        \Dot{\theta} \Delta t \\
        0 \\
        0 \\
        0 \\
    \end{bmatrix}
    \label{eq:ctrv_regular}
\end{equation}
where, $\Delta t$ is the sampling time. Note that in Equation.\eqref{eq:ctrv_regular}, $z$ is assumed to evolve with constant velocity. In the case of zero turning rate (i.e. $\Dot{\theta} = 0$), 

\begin{equation}
    \mathrm{x}_{t+1} = \mathrm{x}_t + \begin{bmatrix}
        \mathtt{v} \cos (\theta) &
        \mathtt{v} \sin (\theta) &
        \Dot{z} \Delta t &
        \Dot{\theta} \Delta t &
        0 &
        0 &
        0 &
    \end{bmatrix}^T
    \label{eq:ctrv_zero_turning_rate}
\end{equation}

The state vector of a pedestrian is

\begin{equation}
    \mathrm{x} = \begin{bmatrix}
        x & y & z & \theta & \Dot{x} & \Dot{y} & \Dot{z} & \Dot{\theta}
    \end{bmatrix}^T
\end{equation}
The motion of pedestrians is predicted according to CV model

\begin{equation}
    \mathrm{x}_{t+1} = \mathrm{x}_t + \begin{bmatrix}
        \Dot{x} &
        \Dot{y} &
        \Dot{z}&
        \Dot{\theta}&
        0 &
        0 &
        0 &
        0
    \end{bmatrix}^T \cdot \Delta t
    \label{eq:cv_model}
\end{equation}

\section{Experiments}
The effectiveness of our method is demonstrated by benchmarking against the SORT-style baseline model on 3 large scale datasets: KITTI, NuScenes, and Waymo. In addition, we perform ablation study using NuScenes dataset to better understand the impact of each component on our system's general performance.

\subsection{Tracking Results}

\textbf{Evaluation Metrics}: Classically, MOT systems are evaluated by the CLEAR MOT metrics \cite{bernardin2008evaluating}. As pointed out by \cite{leal2017tracking} and later by \cite{weng2020ab3dmot}, there is a linear relation between MOTA and object detectors' recall rate, as a result, MOTA does not provide a well-rounded evaluation performance of trackers. To remedy this, \cite{weng2020ab3dmot} proposes to average MOTA and MOTP over a range of recall rate, resulting in two integral metrics AMOTA and AMOTP which become the norm in recent benchmarks.

\textbf{Datasets} To verify the effectiveness of our method, we benchmark it on 3 popular autonomous driving datasets which offer 3D MOT benchmark: KITTI, NuScenes, and Waymo. These datasets are collections of driving sequences collected in various environment using a multi-modal sensor suite including LiDAR. KITTI tracking benchmark interests in two classes of object which are cars and pedestrians. Initially, KITTI tracking was designed for MOT in 2D images and recently \cite{weng2020ab3dmot} adapts it to 3D MOT. NuScenes concerns a larger set of objects which comprises of cars, bicycles, buses, trucks, pedestrians, motorcycles, trailers. Waymo shares the same interest as NuScenes but groups car-like vehicles into meta class.

\textbf{Public Detection}: As can be seen in Table.\ref{tab:agg-tracking-methods}, AMOTA highly depends on the precision of object detectors. Therefore, to have a fair comparison, the baseline detection results made publicly available by the benchmarks are used as the input to our tracking system. Specifically, we use MEGVII detection and PointPillars with PPBA detection for NuScenes and Waymo, respectively.

The performance of our model compared to the SORT-style baseline model in 3 popular benchmarks are shown in Table.\ref{tab:quant-perf}. As can be seen, our model consistently outperforms the baseline model in term of the primary metric AMOTA. The main reason of this is the lower ID switches and trajectory fragments of ours which shows the better ability of establishing track-to-detection correspondence compared to SORT-style algorithm.

\begin{table}[h] 
\caption{Quantitative performance of our model on KITTI, NuScenes, and Waymo validation set.\label{tab:quant-perf}}
\begin{tabular}{ m{1.5cm} m{2cm} m{1.25cm} m{1.25cm} m{0.75cm} m{0.75cm} m{1cm} m{1cm} m{0.75cm} m{0.75cm} } 
\toprule
Dataset & Method & \textbf{AMOTA}$\uparrow$ & AMOTP$\downarrow$ & MT$\uparrow$ & ML$\downarrow$ & FP$\downarrow$ & FN$\downarrow$ & IDS$\downarrow$ & FRAG$\downarrow$ \\
\midrule
\multirow{2}{1cm}{KITTI} 
& Ours & \textbf{0.415} & 0.691 & N/A & N/A & 766 & 3721 & 10 & 259 \\
& AB3DMOT\cite{weng2020ab3dmot} & 0.377 & \textbf{0.648} & N/A & N/A & \textbf{696} & \textbf{3713} & \textbf{1} & \textbf{93} \\
& & & & & & & & & \\
\multirow{2}{1cm}{NuScenes} 
& Ours & \textbf{0.583} & \textbf{0.748} & \textbf{3617} & 1885 & 13439 & \textbf{28119} & \textbf{512} & \textbf{511} \\
& StanfordIPRL-TRI\cite{chiu2020probabilistic} & 0.561 & 0.800 & 3432 & \textbf{1857} & \textbf{12140} & 28387 & 679 & 606 \\
\bottomrule
\end{tabular}
\end{table}

\subsection{Ablation Study}
In this ablation study, default method is the method presented in Section.\ref{sec:method} which has 

\begin{itemize}
    \item Two stages of data association (local and global). Each stage is formulated as a LAP and solved by a greedy matching algorithm \cite{chiu2020probabilistic}.
    \item The affinity function the sum of position affinity and size affinity (as in Equation.\eqref{eq:affinity_general}).
    \item The motion model is Constant Turning Rate and Velocity (CTRV) for car-like objects (cars, buses, trucks, trailers, bicycles) and Constant Veloctiy (CV) for pedestrians.   
\end{itemize}
To understand the effect of each component on the system's general performance, we modify or remove each of them and carry out experiment with the rest of the system being kept the same as the default method and the same hyperparameters. The changes and the resulted performance are shown in Table.\ref{tab:ablation-study}.

\begin{table}[htb]
\caption{Ablation study using NuScenes dataset.\label{tab:ablation-study}}
\begin{tabular}{ m{3cm} m{1.25cm} m{1.25cm} m{0.75cm} m{0.75cm} m{1cm} m{1cm} m{0.75cm} m{0.75cm} } 
\toprule
Method & \textbf{AMOTA}$\uparrow$ & AMOTP$\downarrow$ & MT$\uparrow$ & ML$\downarrow$ & FP$\downarrow$ & FN$\downarrow$ & IDS$\downarrow$ & FRAG$\downarrow$ \\
\midrule
Default & 0.583 & 0.748 & \textbf{3617} & 1885 & 13439 & 28119 & 512 & 511 \\
Hungarian for LAP & \textbf{0.587} & \textbf{0.743} & 3609 & \textbf{1880} & 13667 & \textbf{28070} & 596 & 573 \\ 
No ReID & 0.583 & 0.748 & 3616 & 1882 & 13429 & 28100 & \textbf{504} & \textbf{510} \\
Global assoc only & 0.327 & 0.924 & 2575 & 2244 & 26244 & 38315 & 4215 & 3038 \\
Const Velocity only & 0.567 & 0.781 & 3483 & 1966 & \textbf{12649} & 29427 & 718 & 606 \\
No size affinity & 0.581 & 0.748 & 3595 & 1904 & 13423 & 28448 & 512 & 508 \\
\bottomrule
\end{tabular}
\end{table}


\bibliographystyle{unsrt}  
\bibliography{references}  

\begin{thebibliography}{10}

\bibitem{ren2016faster}
Shaoqing Ren, Kaiming He, Ross Girshick, and Jian Sun.
\newblock Faster r-cnn: Towards real-time object detection with region proposal
  networks.
\newblock {\em IEEE transactions on pattern analysis and machine intelligence},
  39(6):1137--1149, 2016.

\bibitem{liu2016ssd}
Wei Liu, Dragomir Anguelov, Dumitru Erhan, Christian Szegedy, Scott Reed,
  Cheng-Yang Fu, and Alexander~C Berg.
\newblock Ssd: Single shot multibox detector.
\newblock In {\em European conference on computer vision}, pages 21--37.
  Springer, 2016.

\bibitem{redmon2016you}
Joseph Redmon, Santosh Divvala, Ross Girshick, and Ali Farhadi.
\newblock You only look once: Unified, real-time object detection.
\newblock In {\em Proceedings of the IEEE conference on computer vision and
  pattern recognition}, pages 779--788, 2016.

\bibitem{bewley2016simple}
Alex Bewley, Zongyuan Ge, Lionel Ott, Fabio Ramos, and Ben Upcroft.
\newblock Simple online and realtime tracking.
\newblock In {\em 2016 IEEE International Conference on Image Processing
  (ICIP)}, pages 3464--3468. IEEE, 2016.

\bibitem{scheidegger2018mono}
Samuel Scheidegger, Joachim Benjaminsson, Emil Rosenberg, Amrit Krishnan, and
  Karl Granstr{\"o}m.
\newblock Mono-camera 3d multi-object tracking using deep learning detections
  and pmbm filtering.
\newblock In {\em 2018 IEEE Intelligent Vehicles Symposium (IV)}, pages
  433--440. IEEE, 2018.

\bibitem{weng2020ab3dmot}
Xinshuo Weng, Jianren Wang, David Held, and Kris Kitani.
\newblock Ab3dmot: A baseline for 3d multi-object tracking and new evaluation
  metrics.
\newblock {\em arXiv preprint arXiv:2008.08063}, 2020.

\bibitem{kuhn1955hungarian}
Harold~W Kuhn.
\newblock The hungarian method for the assignment problem.
\newblock {\em Naval research logistics quarterly}, 2(1-2):83--97, 1955.

\bibitem{chiu2020probabilistic}
Hsu kuang Chiu, Antonio Prioletti, Jie Li, and Jeannette Bohg.
\newblock Probabilistic 3d multi-object tracking for autonomous driving, 2020.

\bibitem{zhou2019iou}
Dingfu Zhou, Jin Fang, Xibin Song, Chenye Guan, Junbo Yin, Yuchao Dai, and
  Ruigang Yang.
\newblock Iou loss for 2d/3d object detection.
\newblock In {\em 2019 International Conference on 3D Vision (3DV)}, pages
  85--94. IEEE, 2019.

\bibitem{liang2020pnpnet}
Ming Liang, Bin Yang, Wenyuan Zeng, Yun Chen, Rui Hu, Sergio Casas, and Raquel
  Urtasun.
\newblock Pnpnet: End-to-end perception and prediction with tracking in the
  loop.
\newblock In {\em Proceedings of the IEEE/CVF Conference on Computer Vision and
  Pattern Recognition}, pages 11553--11562, 2020.

\bibitem{yin2020center}
Tianwei Yin, Xingyi Zhou, and Philipp Kr{\"a}henb{\"u}hl.
\newblock Center-based 3d object detection and tracking.
\newblock {\em arXiv preprint arXiv:2006.11275}, 2020.

\bibitem{luo2018fast}
Wenjie Luo, Bin Yang, and Raquel Urtasun.
\newblock Fast and furious: Real time end-to-end 3d detection, tracking and
  motion forecasting with a single convolutional net.
\newblock In {\em Proceedings of the IEEE conference on Computer Vision and
  Pattern Recognition}, pages 3569--3577, 2018.

\bibitem{geiger2013vision}
Andreas Geiger, Philip Lenz, Christoph Stiller, and Raquel Urtasun.
\newblock Vision meets robotics: The kitti dataset.
\newblock {\em The International Journal of Robotics Research},
  32(11):1231--1237, 2013.

\bibitem{caesar2020nuscenes}
Holger Caesar, Varun Bankiti, Alex~H Lang, Sourabh Vora, Venice~Erin Liong,
  Qiang Xu, Anush Krishnan, Yu~Pan, Giancarlo Baldan, and Oscar Beijbom.
\newblock nuscenes: A multimodal dataset for autonomous driving.
\newblock In {\em Proceedings of the IEEE/CVF Conference on Computer Vision and
  Pattern Recognition}, pages 11621--11631, 2020.

\bibitem{sun2020scalability}
Pei Sun, Henrik Kretzschmar, Xerxes Dotiwalla, Aurelien Chouard, Vijaysai
  Patnaik, Paul Tsui, James Guo, Yin Zhou, Yuning Chai, Benjamin Caine, et~al.
\newblock Scalability in perception for autonomous driving: Waymo open dataset.
\newblock In {\em Proceedings of the IEEE/CVF Conference on Computer Vision and
  Pattern Recognition}, pages 2446--2454, 2020.

\bibitem{pmbm}
Angel~F. Garcia-Fernandez, Jason~L. Williams, Karl Granstrom, and Lennart
  Svensson.
\newblock Poisson multi-bernoulli mixture filter: Direct derivation and
  implementation.
\newblock {\em IEEE Transactions on Aerospace and Electronic Systems},
  54(4):1883–1901, Aug 2018.

\bibitem{zhu2019megvii}
Benjin Zhu, Zhengkai Jiang, Xiangxin Zhou, Zeming Li, and Gang Yu.
\newblock Class-balanced grouping and sampling for point cloud 3d object
  detection.
\newblock {\em arXiv preprint arXiv:1908.09492}, 2019.

\bibitem{zhou2019objects}
Xingyi Zhou, Dequan Wang, and Philipp Krähenbühl.
\newblock Objects as points, 2019.

\bibitem{ding20201st}
Zhuangzhuang Ding, Yihan Hu, Runzhou Ge, Li~Huang, Sijia Chen, Yu~Wang, and Jie
  Liao.
\newblock 1st place solution for waymo open dataset challenge -- 3d detection
  and domain adaptation, 2020.

\bibitem{ge2020afdet}
Runzhou Ge, Zhuangzhuang Ding, Yihan Hu, Yu~Wang, Sijia Chen, Li~Huang, and
  Yuan Li.
\newblock Afdet: Anchor free one stage 3d object detection, 2020.

\bibitem{shi2020pv}
Shaoshuai Shi, Chaoxu Guo, Li~Jiang, Zhe Wang, Jianping Shi, Xiaogang Wang, and
  Hongsheng Li.
\newblock Pv-rcnn: Point-voxel feature set abstraction for 3d object detection.
\newblock In {\em Proceedings of the IEEE/CVF Conference on Computer Vision and
  Pattern Recognition}, pages 10529--10538, 2020.

\bibitem{cheng2020improving}
Shuyang Cheng, Zhaoqi Leng, Ekin~Dogus Cubuk, Barret Zoph, Chunyan Bai, Jiquan
  Ngiam, Yang Song, Benjamin Caine, Vijay Vasudevan, Congcong Li, et~al.
\newblock Improving 3d object detection through progressive population based
  augmentation.
\newblock {\em arXiv preprint arXiv:2004.00831}, 2020.

\bibitem{bae2014robust}
Seung-Hwan Bae and Kuk-Jin Yoon.
\newblock Robust online multi-object tracking based on tracklet confidence and
  online discriminative appearance learning.
\newblock In {\em Proceedings of the IEEE conference on computer vision and
  pattern recognition}, pages 1218--1225, 2014.

\bibitem{arnold2019survey}
Eduardo Arnold, Omar~Y Al-Jarrah, Mehrdad Dianati, Saber Fallah, David Oxtoby,
  and Alex Mouzakitis.
\newblock A survey on 3d object detection methods for autonomous driving
  applications.
\newblock {\em IEEE Transactions on Intelligent Transportation Systems},
  20(10):3782--3795, 2019.

\bibitem{geiger2013traffic}
Andreas Geiger, Martin Lauer, Christian Wojek, Christoph Stiller, and Raquel
  Urtasun.
\newblock 3d traffic scene understanding from movable platforms.
\newblock {\em IEEE transactions on pattern analysis and machine intelligence},
  36(5):1012--1025, 2013.

\bibitem{leal2015motchallenge}
Laura Leal-Taix{\'e}, Anton Milan, Ian Reid, Stefan Roth, and Konrad Schindler.
\newblock Motchallenge 2015: Towards a benchmark for multi-target tracking.
\newblock {\em arXiv preprint arXiv:1504.01942}, 2015.

\bibitem{mauri2020deep}
Antoine Mauri, Redouane Khemmar, Benoit Decoux, Nicolas Ragot, Romain Rossi,
  Rim Trabelsi, R{\'e}mi Boutteau, Jean-Yves Ertaud, and Xavier Savatier.
\newblock Deep learning for real-time 3d multi-object detection, localisation,
  and tracking: Application to smart mobility.
\newblock {\em Sensors}, 20(2):532, 2020.

\bibitem{bae2017confidence}
Seung-Hwan Bae and Kuk-Jin Yoon.
\newblock Confidence-based data association and discriminative deep appearance
  learning for robust online multi-object tracking.
\newblock {\em IEEE transactions on pattern analysis and machine intelligence},
  40(3):595--610, 2017.

\bibitem{yang2019efficient}
Honghong Yang, Jinming Wen, Xiaojun Wu, Li~He, and Shahid Mumtaz.
\newblock An efficient edge artificial intelligence multipedestrian tracking
  method with rank constraint.
\newblock {\em IEEE Transactions on Industrial Informatics}, 15(7):4178--4188,
  2019.

\bibitem{bernardin2008evaluating}
Keni Bernardin and Rainer Stiefelhagen.
\newblock Evaluating multiple object tracking performance: the clear mot
  metrics.
\newblock {\em EURASIP Journal on Image and Video Processing}, 2008:1--10,
  2008.

\bibitem{leal2017tracking}
Laura Leal-Taix{\'e}, Anton Milan, Konrad Schindler, Daniel Cremers, Ian Reid,
  and Stefan Roth.
\newblock Tracking the trackers: an analysis of the state of the art in
  multiple object tracking.
\newblock {\em arXiv preprint arXiv:1704.02781}, 2017.

\end{thebibliography}

\end{document}